\pdfoutput=1

\documentclass[11pt]{article}

\usepackage[]{acl}

\usepackage{times}
\usepackage{latexsym}
\usepackage{microtype}
\usepackage[T1]{fontenc}
\usepackage{tgbonum}
\usepackage{amsmath}
\usepackage{multirow}
\usepackage{hyperref}
\usepackage{enumitem}
\usepackage{caption}
\usepackage{graphicx}
\usepackage{subfigure}
 \usepackage{float} 
\usepackage[linesnumbered,ruled,vlined]{algorithm2e}
\usepackage[T1]{fontenc}

\usepackage[utf8]{inputenc}

\usepackage{microtype}

\title{Dict-NMT: Bilingual Dictionary based NMT for Extremely Low Resource Languages}

\author{Nalin Kumar$^{1}$  , Deepak Kumar$^{2}$ , Subhankar Mishra$^{3}$ \\
  School of Computer Sciences, NISER, Bhubaneswar- 752050 \\
  Homi Bhabha National Institute, Mumbai-400094, India \\
  \texttt{\{nalin.kumar$^{1}$, deepak.kumar$^{2}$, smishra$^{3}$\}@niser.ac.in} \\}

\begin{document}
\maketitle
\begin{abstract}
Neural Machine Translation (NMT) models have been effective on large bilingual datasets. However, the existing methods and techniques show that the model's performance is highly dependent on the number of examples in training data. For many languages, having such an amount of corpora is a far-fetched dream. Taking inspiration from monolingual speakers exploring new languages using bilingual dictionaries, we investigate the applicability of bilingual dictionaries for languages with extremely low, or no bilingual corpus. In this paper, we explore methods using bilingual dictionaries with an NMT model to improve translations for extremely low resource languages. We extend this work to multilingual systems, exhibiting zero-shot properties. We present a detailed analysis of the effects of the quality of dictionaries, training dataset size, language family, etc., on the translation quality. Results on multiple low-resource test languages show a clear advantage of our bilingual dictionary-based method over the baselines.

\end{abstract}

\section{Introduction}
With the growing interest in improving automatic translation systems, deep learning-based models have played a significant role. They have a ubiquitous influence on such solutions. Neural Machine translation has been ruling the roost in recent times, both in academia and industries. It has outperformed other translation methods, and even human translators for some languages \cite{bojar-etal-2016-findings} \cite{bentivogli2016neural} \cite{barrault-etal-2020-findings}. The encoder-decoder framework of NMT models allows them to transfer the semantic and syntactic information more precisely. \par 
One of the major challenges for such languages is training corpora of sufficient size. Such models need large bilingual or monolingual datasets, which usually range between $1$-$50$ million parallel sentences. For the extremely low resourced languages, datasets smaller than $20$ thousand parallel sentences, NMT models have not been that successful \cite{ostling2017neural}. The standard approach to this problem has mostly relied on techniques such as transfer learning \cite{zoph2016transfer}, and data augmentation approaches such as back-translation \cite{sennrich2015improving} \cite{przystupa2019neural} and data diversification \cite{nguyen2019data}. \par
The use of prior knowledge sources for translation of low-resource languages, such as bilingual dictionaries, is still under-explored. The work of \cite{duan2020bilingual} and \cite{nag2020incorporating} explore the use of a bilingual dictionary, but they use an additional large monolingual corpus. In contrast, we use an extremely small test language's bilingual corpus or no bilingual corpora of the test language at all. The existing approaches mainly depend on the availability of additional corpora like target monolingual corpus and target-to-source model for back-translation, sister language for transfer learning, or additional computations as in data diversification. One of the most common and widely available prior knowledge resources across low-resourced languages is the bilingual dictionary which has shown potential in NMT in recent times. \cite{pourdamghani2019translating} explores the possibility of utilizing a bilingual dictionary for unsupervised translation. Our method is highly inspired by their two-step approach. However, we provide extensive experiments on the correlation between the quality of dictionaries, dataset size, and BLEU.\par
\begin{figure}[H]
\centering
\includegraphics[width=6cm, height=4cm]{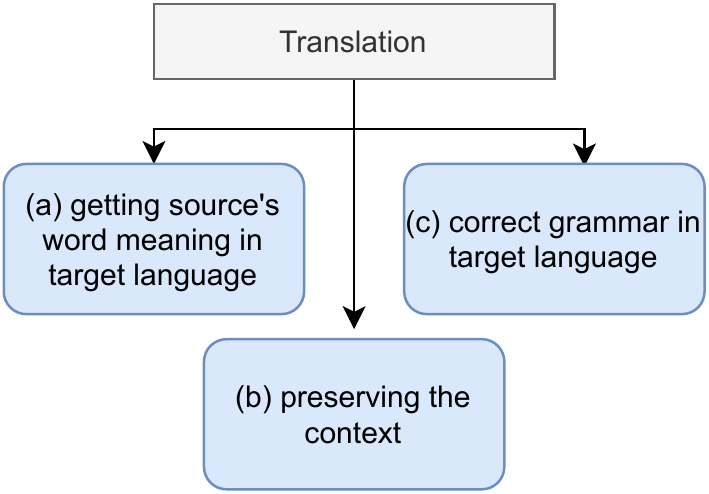}
\caption{Translation (a) word mapping task, which can be partially, or completely achieved with bilingual dictionary lookup. (b) is about the association a word has with its surrounding words, which in turn affects its alignment. (c) is the transformation of syntactic features of the source language to the source language. }
\label{fig:translation_points}
\end{figure}

In our work, we explore using a bilingual dictionary to translate extremely low languages. Any meaningful translation requires us to address the points as illustrated in Figure \ref{fig:translation_points}. In extremely low resource languages, NMT falls behind given the lack of enormous quantity of data required to train them properly. We study the potential of assisting the NMT models with the contextual dictionary transformation. Our proposed method involves the
use of bilingual dictionary for addressing the points (Figure \ref{fig:translation_points} (a) and (b)) and an NMT model for (Figure \ref{fig:translation_points} (c)), i.e., we use an NMT model to transform a distorted sentence into a meaningful sentence within the same language. Using this method, we propose two simple frameworks which can be extremely useful for languages with extremely less or no corpus available but having a bilingual dictionary. Summarizing the contributions of our paper as follows: 
\begin{itemize}
  \item We introduce a simple and effective method for incorporating a bilingual dictionary in a neural machine translation task.
  \item We propose a one-to-one bilingual dictionary based NMT model for extremely low-resource languages.
  \item We propose a many-to-one NMT model capable of translating for languages it has never seen in the training sets.
\end{itemize} 



We provide a brief description of our method in Section $2$. We discuss the usage of the bilingual dictionary, tokenizer, and the NMT model. We explore the applicability of our proposed method in two settings, extremely less corpus and no corpus available for concerned language. In Section $3$, we describe our one-to-one translation framework useful for translation in extremely low resource setting. We provide a detailed analysis with comparison among translation quality, dataset size and dictionary quality. In Section $4$, we provide detailed information about our proposed many-to-one translation framework, which shows zero-shot property. We summarize and conclude our results and contributions in Section $5$.

\section{Dict-NMT: Assisting NMT model with bilingual dictionary}
\begin{figure}[]
\centering
\includegraphics[width=4.7cm, height=5.6cm]{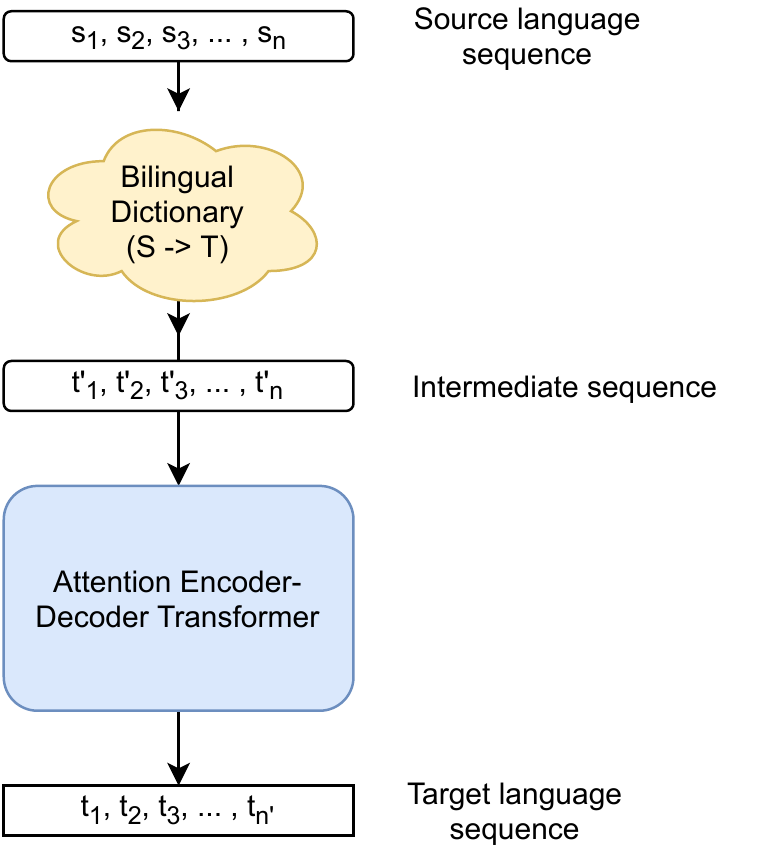}
\caption{Our proposed method involving bilingual dictionary for NMT.}
\label{fig:model_figure}
\end{figure}

We propose a simple yet effective method of translation, dict-NMT, using an NMT model with the help of the respective languages' bilingual dictionaries. We use a bilingual dictionary as a word-to-word translator to convert words from the source language to an intermediate sequence. This distorted sentence in the target language is then fed to an NMT model, here Transformer, to learn the relation between the intermediate sequence and ground truth (Figure \ref{fig:model_figure}). This opens up doors for various frameworks for translation. A straightforward way is to apply this method to a one-to-one translation system (Section $3$).
Furthermore, one can also devise a many-to-one translation framework (Section $4$), where the NMT model is trained on word-to-word translations from various languages. This generalized model can then be used even for languages that were not used in the training data. Other possible ways include fine-tuning the generalized model on a specific language. Other data augmentation methods, such as backtranslation and data diversification, are also applicable to our proposed method. Another possible way of augmenting data is by adding intra-shuffled (i.e., words within a sentence $s$ are shuffled), noisy (replace tokens in $s$ with random tokens with some probability) sentences from the target language to the training data. We leave these methods for future work.

\subsection{Bilingual Dictionary}
A dictionary is a map of words from the source language to the target language, where the mapping can be one to many. Here, we consider mappings that are word to word and not word to phrase. 

First, we change the source language sentences into an intermediate sequence using the dictionary. This step would reduce the workload on our NMT model from learning the word meanings from the available small dataset. If a word in the source sentence is present in the dictionary, then it is converted accordingly in the intermediate sequence; otherwise, it remains unchanged in the intermediate sequence, i.e., we consider the word to be in the target language space. When using the dictionary, multiple target language words might exist as meaning for a source language word. We settle this problem of polysemy by selecting the word most similar to the previous word's dictionary translation (using the target language's pre-trained word embeddings). This would help us to preserve contextual information. \par
More precisely, for any source language $S$ and target language $T$ with a bilingual dictionary $D_{S \rightarrow T}$, the first step is to translate the text in $S$ word-to-word to $T$ using $D_{S \rightarrow T}$. If the mapping is not available for any word $w$ in $S$, it is mapped to itself and is considered a random noise in $T$. 
For the case of polysemous words, we take the help of word embeddings of $T$. We select the word (in $T$) most similar to the previous word's dictionary translation (using target language's pre-trained word embeddings). For instance, given is a sentence $s = \{s_{1}, s_{2}, ..., s_{n}\}$ in $S$ with word-to-word translation $t = \{t_{1}, t_{2}, ..., t_{n}\}$ in $T$. For any $s_{i}(i > 1)$ having dictionary translations $D_{S \rightarrow T}(s_{i}) = \{t_{i}^{1}, t_{i}^{2}, ..., t_{i}^{m}\}$, we select its translation as 
\[t_{i} = \underset{t \in D_{S \rightarrow T}(s_{i})}{argmax} similarity(t_{j}, t)\]
where $j = \max\{j' < i | s_{j} \in Dom(D_{S \rightarrow T})\}$. We randomly select translation for the first occurring polysemous word.  Here, for our experiments, we assume that the target language is a popular one, thus, decent word embeddings for $T$ exist. This method would help us to preserve the contextual information. However, if the first randomly selected word is erroneous, the trailing polysemous words might have incorrect translations.
\subsection{Tokenizer}
Tokenization is the process of breaking the given text into smaller chunks. Since the model input and output are in the same language, we share the tokenizer for both of them. The intermediate representation might consist of words from a foreign language. Thus, instead of using the traditional whitespace tokenizer and giving all such words a <OOV> token, we use subspace tokenizer to handle the large amount of out-of-vocabulary words. This way, the noise created by the tokens of foreign language would help the model be more robust.

\begin{algorithm*}
  \caption{\small Dict-NMT. $D_{i} = (S_{i}, T)$ is a set of parallel sentences from language $S_{i}$ to $T$. Corresponding to the language pair $(S_{i}, T)$, we have a bilingual dictionary $B_{i}$ where $B_{i}(s)$ is a word-to-word translation of $s$. We train the model $M$ on $\{D_{i}\}^{n}_{i=1}$ using $\{B_{i}\}^{n}_{i=1}$. }\label{algo}
  
    \DontPrintSemicolon
    
    \SetKwFunction{FMain}{Train}
    \SetKwProg{Fn}{Procedure}{:}{}
    \Fn{\FMain{$\{D_{i}\}^{n}_{i=1}$, $\{B_{i}\}^{n}_{i=1}$, p}}{
          M \textit{randomly initialised NMT model}.\;
           Train M on {\fontfamily{qcr}\selectfont create$\_$dataset}($\{D_{i}\}^{n}_{i=1}$, $\{B_{i}\}^{n}_{i=1}$, p) until it converges\;
          \KwRet M\;
    }
  
    \SetKwFunction{FMain}{create$\_$dataset}
    \SetKwProg{Fn}{Function}{$:$}{}
    \Fn{\FMain{$\{D_{i}\}^{n}_{i=1}$, $\{B_{i}\}^{n}_{i=1}$, p }}
    {
          $D' = \phi$\;
          \For{D $\in$ $\{D_{i}\}^{n}_{i=1}$}{\For{(s, t) $\in$ $D$}{\If{$count_{T}$($B_{i}(s)$) $\geq$ $p$}{$D'$ = $D'$ $\cup$ ($B_{i}(s)$, t)}}}\;
          \tcc{\small $count_T(B_{s}$) = $\%$ of words in $s$ having dictionary translations}\;
          \textit{shuffle} $D'$ \;
          \KwRet $D'$\; }
  \end{algorithm*}

\subsection{NMT Model}
Since both intermediate sequence and target belong to the same language, the NMT model is relieved from learning the word meanings. The model will now try to focus primarily on learning the grammar for the target language space. The NMT model learns the mappings from the source invariant representations from various languages to the target language and tries to generalize which would benefit unknown languages. \par
Our proposed method can be applied to any NMT model. For our experiments, we use the state-of-the-art Transformer \cite{vaswani2017attention} model. Since the intermediate sentence, i.e., the input for the Transformer, in itself does not make any sense, the attention mechanism helps to understand the dependencies of words through the whole sequence. The encoder-decoder framework allows us to find the meaning of the words not translated by the dictionary while preserving the context.



\section{Dict-NMT for one-to-one translation}
We propose a dictionary-based one-to-one translation framework for extremely low resource settings. Given a language pair $(S, T)$, we train an NMT model on the word-to-word dictionary translations of $S$, and T $(i=1$ in Algo $\ref{algo})$.
\subsection{Experimental Settings}
We extensively check the effectiveness of the dictionary (by varying the dictionary percentage) across five European languages' translation tasks and the size of the bilingual corpora. We keep Transformer as our baseline model.
We use $4$ layer Transformer with $100$ embedding/hidden units and $400$ feed-forward filter size. We tie source and target embeddings. We keep batch size $32$, epochs $50$, dropout $0.1$ and optimizer Adam. In this work, we use the pre-trained BERT WordPiece tokenizer \cite{devlin-etal-2019-bert}, a subword tokenizer.\\
\subsubsection{Bilingual dictionary}
For our experiments, we use the publicly available Facebook MUSE's\footnote[1]{https://github.com/facebookresearch/MUSE} bilingual dictionary, which consists of 110 large-scale ground-truth bilingual dictionaries \cite{conneau2017word}. 
For preserving the context while dictionary translation, we use the Fasttext embeddings \cite{bojanowski2017enriching}.
\subsubsection{Data}
For our experiments, we consider Europarl v7 parallel corpus \cite{koehn02} for Pt-En, Sv-En, Nl-En, Pt-En, and Fr-En language pairs. Here we selected English as our target language in all the cases. The intuition behind this is that any bilingual dictionary for an extremely low resource language would be created by taking a commonly used language to be of practical use. As English is one such language, we tried our experiments with it. \par

We filter each sentence such that it contains at most $80$ tokens. We use these data in a low resource setting, i.e., only $2$K, $8$K, $16$K, and $20$K data size for each language. We create these datasets according to the percentage of words from each sentence available in the corresponding dictionary, and precisely we did this for $50\%$-$80\%$ (\hyperref[table:1]{Table $1$}). For each data size, $0.05\%$ is the test set, and the rest we use as our training set.

\begin{table}[h]
 \begin{subtable}
 \centering
 \resizebox{\columnwidth}{!}{%
\begin{tabular}{r|rrrrr}
\multicolumn{1}{c|}{\begin{tabular}[c]{@{}c@{}}Dict \\ \%\end{tabular}} &
  \multicolumn{1}{c}{\begin{tabular}[c]{@{}c@{}}Ro-En\\ ($399.37$K)\end{tabular}} &
  \multicolumn{1}{c}{\begin{tabular}[c]{@{}c@{}}Pt-En\\ ($1.96$M)\end{tabular}} &
  \multicolumn{1}{c}{\begin{tabular}[c]{@{}c@{}}Fr-En\\ ($2$M)\end{tabular}} &
  \multicolumn{1}{c}{\begin{tabular}[c]{@{}c@{}}It-En\\ ($1.9$M)\end{tabular}}    &
  \multicolumn{1}{c}{\begin{tabular}[c]{@{}c@{}}Es-En\\ ($1.97$M)\end{tabular}}\\ \hline
$50$ & $104$K & $438.6$K  & $1.4$M & $941$K  & $1.5$M \\
$60$ & $22.6$K  & $77.7$K  & $536.4$K & $202.6$K & $631.6$K\\
$70$ & $4$K  & $11.3$K  & $106.2$K & $26.8$K & $111.4$K\\
$80$ & $975$   & $2.5$K  & $17.7$K  & $4.8$K & $15.9$K\\
$90$ & $244$ & $937$  & $2.6$K  & $1.9$K & $2.5$K \\
$100$ & $230$  & $920$  & $2$K   & $1.8$K & $2$K
\end{tabular}
}
\caption*{Italic}
 \end{subtable}

 \begin{subtable}
 \centering
 \resizebox{\columnwidth}{!}{%
\begin{tabular}{r|rrrr}
\multicolumn{1}{c|}{\begin{tabular}[c]{@{}c@{}}Dict \\ \%\end{tabular}} &
  \multicolumn{1}{c}{\begin{tabular}[c]{@{}c@{}}Da-En\\ ($1.97$M)\end{tabular}} &
  \multicolumn{1}{c}{\begin{tabular}[c]{@{}c@{}}Sv-En\\ ($1.86$M)\end{tabular}} &
  \multicolumn{1}{c}{\begin{tabular}[c]{@{}c@{}}De-En\\ ($1.92$M)\end{tabular}} &
  \multicolumn{1}{c}{\begin{tabular}[c]{@{}c@{}}Nl-En\\ ($2$M)\end{tabular}} \\ \hline
$50$ & $1.3$M & $1.7$M  & $1.9$M & $1.7$M   \\
$60$ & $506.6$K  & $1.3$M  & $1.7$M & $1$M \\
$70$ & $107.9$K  & $563.7$K  & $1$M & $31.7$K \\
$80$ & $20.2$K   & $124.9$K  & $317.5$K  & $55$K \\
$90$ & $2.6$K & $9.2$K  & $28.2$K  & $4.7$K  \\
$100$ & $1.9$K  & $1.8$K  & $3.3$K   & $1.9$K 
\end{tabular}
}
\caption*{Germanic}
 \end{subtable}

  \begin{subtable}
  \centering
  \resizebox{\columnwidth}{!}{%
\begin{tabular}{r|rrrrr}
  \multicolumn{1}{c|}{\begin{tabular}[c]{@{}c@{}}Dict \\ \%\end{tabular}} &
  \multicolumn{1}{c}{\begin{tabular}[c]{@{}c@{}}Bg-En\\ ($406.9	$K)\end{tabular}} &
  \multicolumn{1}{c}{\begin{tabular}[c]{@{}c@{}}Cz-En\\ ($646.6$K)\end{tabular}} &
  \multicolumn{1}{c}{\begin{tabular}[c]{@{}c@{}}Pl-En\\ ($632.57$K)\end{tabular}} &
  \multicolumn{1}{c}{\begin{tabular}[c]{@{}c@{}}Sl-En\\ ($623.49$M)\end{tabular}}    &
  \multicolumn{1}{c}{\begin{tabular}[c]{@{}c@{}}Sk-En\\ ($640.72$K)\end{tabular}}\\ \hline
$50$ & $33.2$K & $136.8$K  & $137.8$K & $64.6$K & $180.4$K  \\
$60$ & $6.4$K  & $36.6$K  & $36.5$K & $16$K & $49.2$K \\
$70$ & $1$K    & $9.5$K  & $8.6$K & $3.9$K & $11.3$K \\
$80$ & $233$K  & $3.1$K  & $2.7$K  & $1.6$K & $3.5$K \\
$90$ & $58$K   & $1.4$K  &  $1.1$K & $1.1$K  & $1.4$K \\
$100$ & $57$K  & $1.3$K  & $1$K   & $1.1$K & $1.3$K 

\end{tabular}
}
\caption*{Slavic}
\end{subtable}
\caption{Dataset size after filtering sentences containing at least Dict\% of dictionary words.}
\label{table:1}
\end{table}

\subsection{Results and Analysis}
We perform intensive experiments on the effectiveness of training data size and dictionary coverage on the performance of the translation system. 
\begin{table}[h!]
\centering
\begin{tabular}{lrrr}
\hline
\multirow{2}{*}{Language} & \multicolumn{1}{c}{Dict \%}                & \multicolumn{2}{c}{BLEU}              \\ \cline{3-4} 
                          & & Base & w D \\ \cline{1-2}
Pt-En       & $55\%$ &  ${6.9}$  & $\mathbf{15.4}$ \\
Sv-En       & $70\%$ & ${8.4}$   & $\mathbf{17.0}$\\
Nl-En     & $65\%$ &  ${5.9}$       & $\mathbf{10.7}$\\
Fr-En       & $65\%$ &  ${8.6}$ & $\mathbf{15.4}$ \\
Da-En      & $65\%$ & ${7.7}$   &  $\mathbf{16.1}$ \\ \hline
\end{tabular}
\caption{Results: Best BLEU score for each language. w D is ``with dictionary", i.e. our proposed method, Base is ``dictionary baseline" which is simply word-to-word translation . We get the best scores for maximum data size (i.e., $20$K). Training set dictionary coverage (Dict $\%$) is given for each corresponding score. The BLEU scores are calculated using SacreBLEU's corpus\_bleu \cite{post-2018-call}}
\label{table:results}
\end{table}
\begin{figure*}[htbp]
\centering
\subfigure[Danish]{\label{}\includegraphics[width=0.32\linewidth]{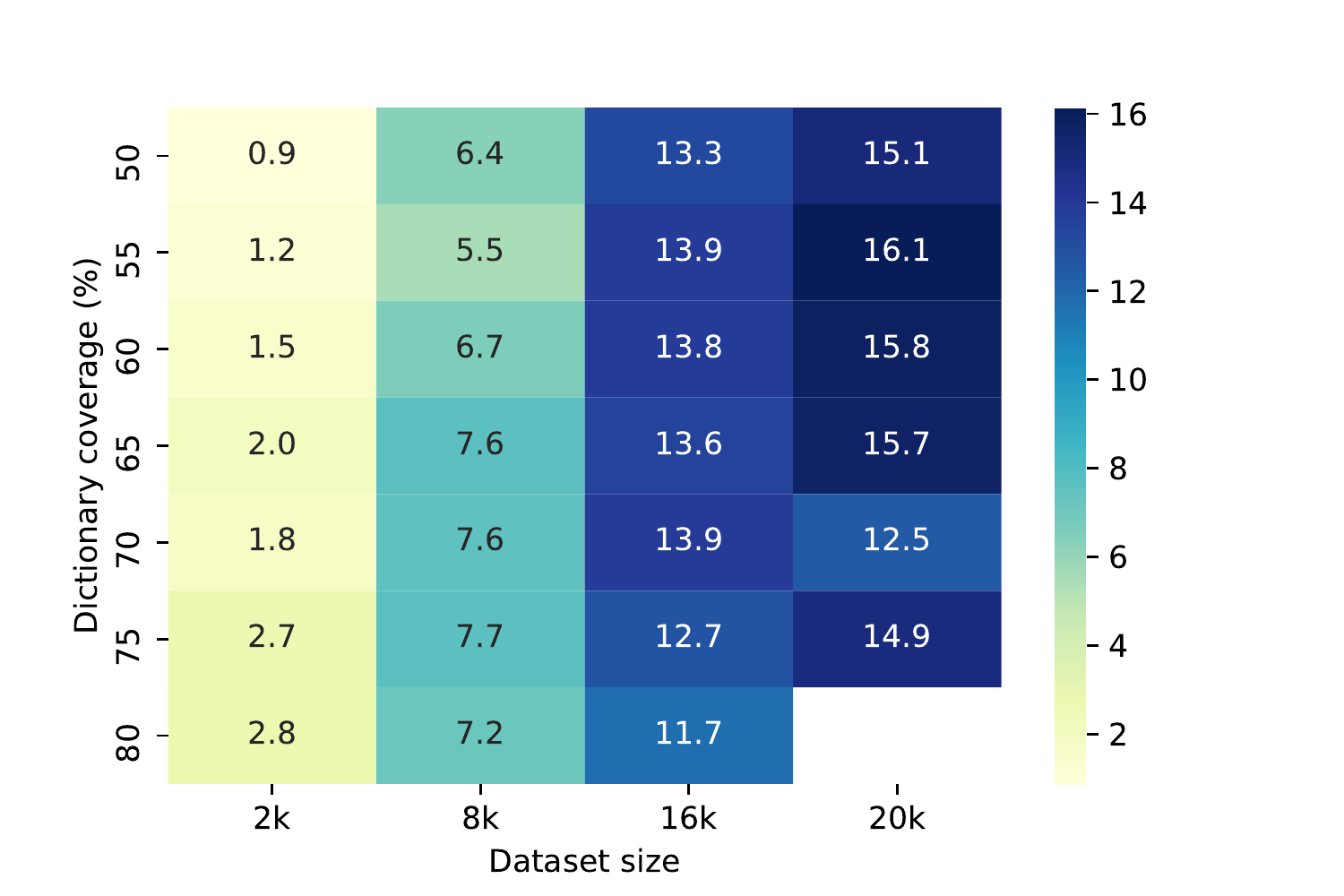}}
\subfigure[Dutch]{\label{}\includegraphics[width=0.32\linewidth]{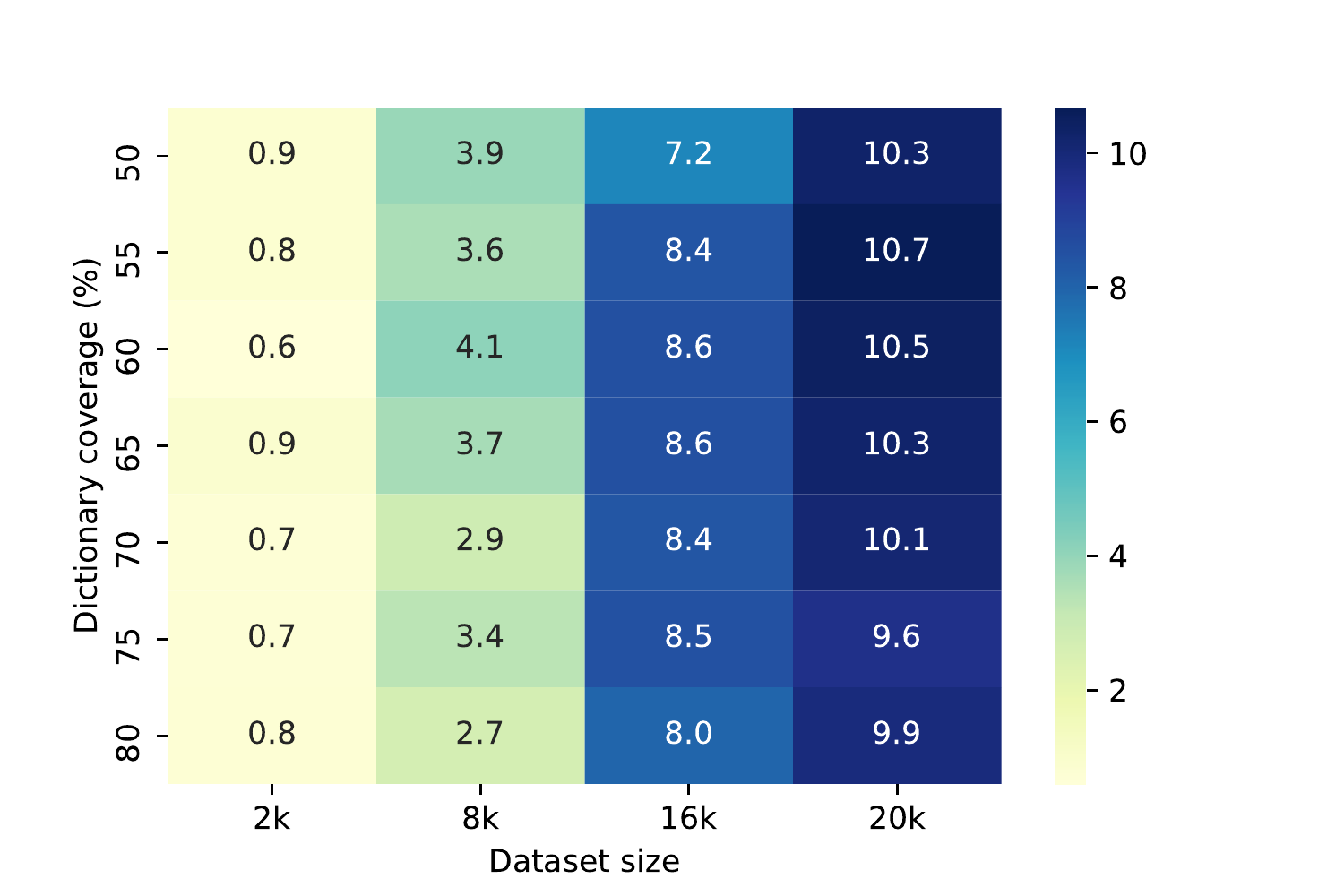}}
\subfigure[French]{\label{}\includegraphics[width=0.32\linewidth]{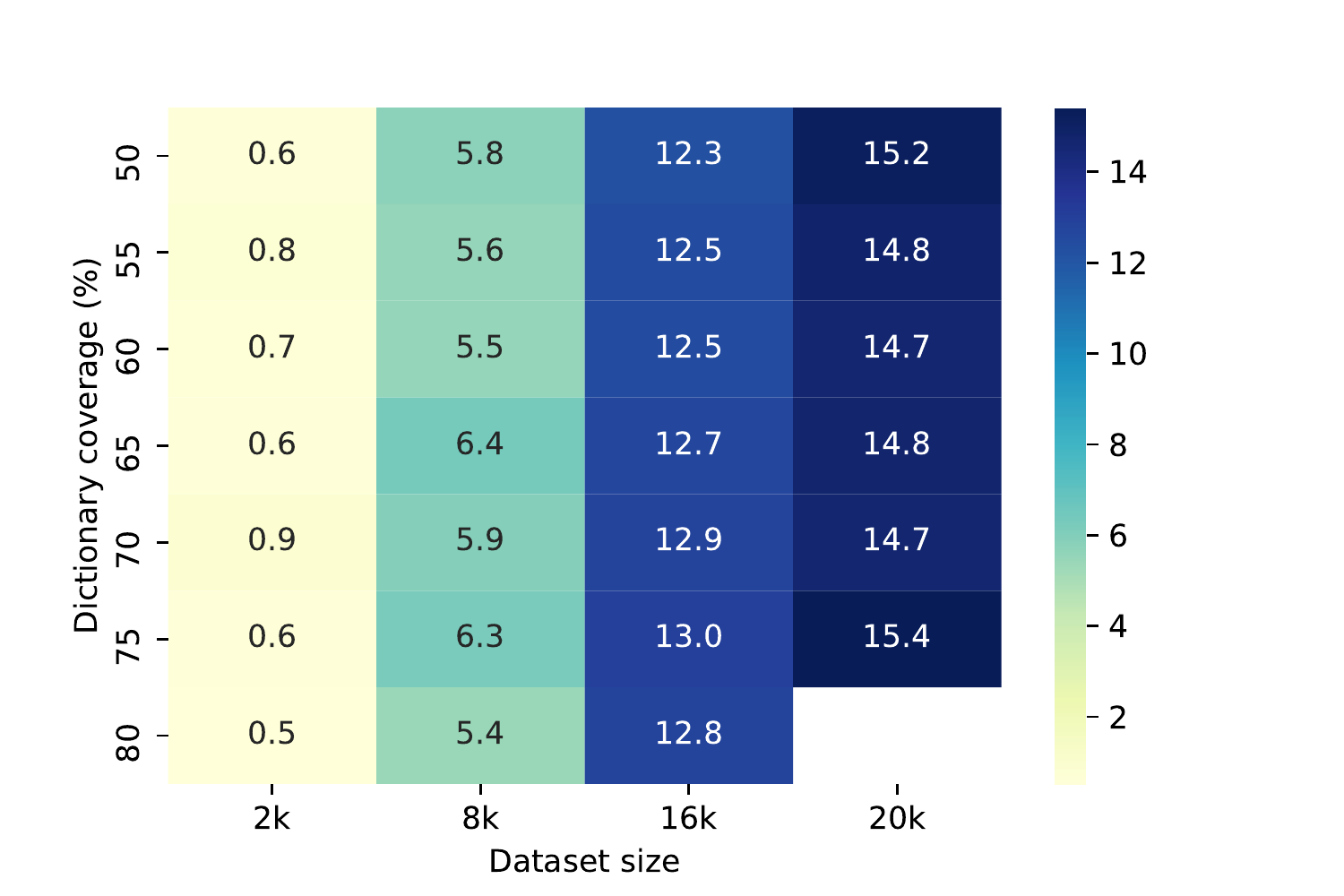}}
\subfigure[Portuguese]{\label{}\includegraphics[width=0.32\linewidth]{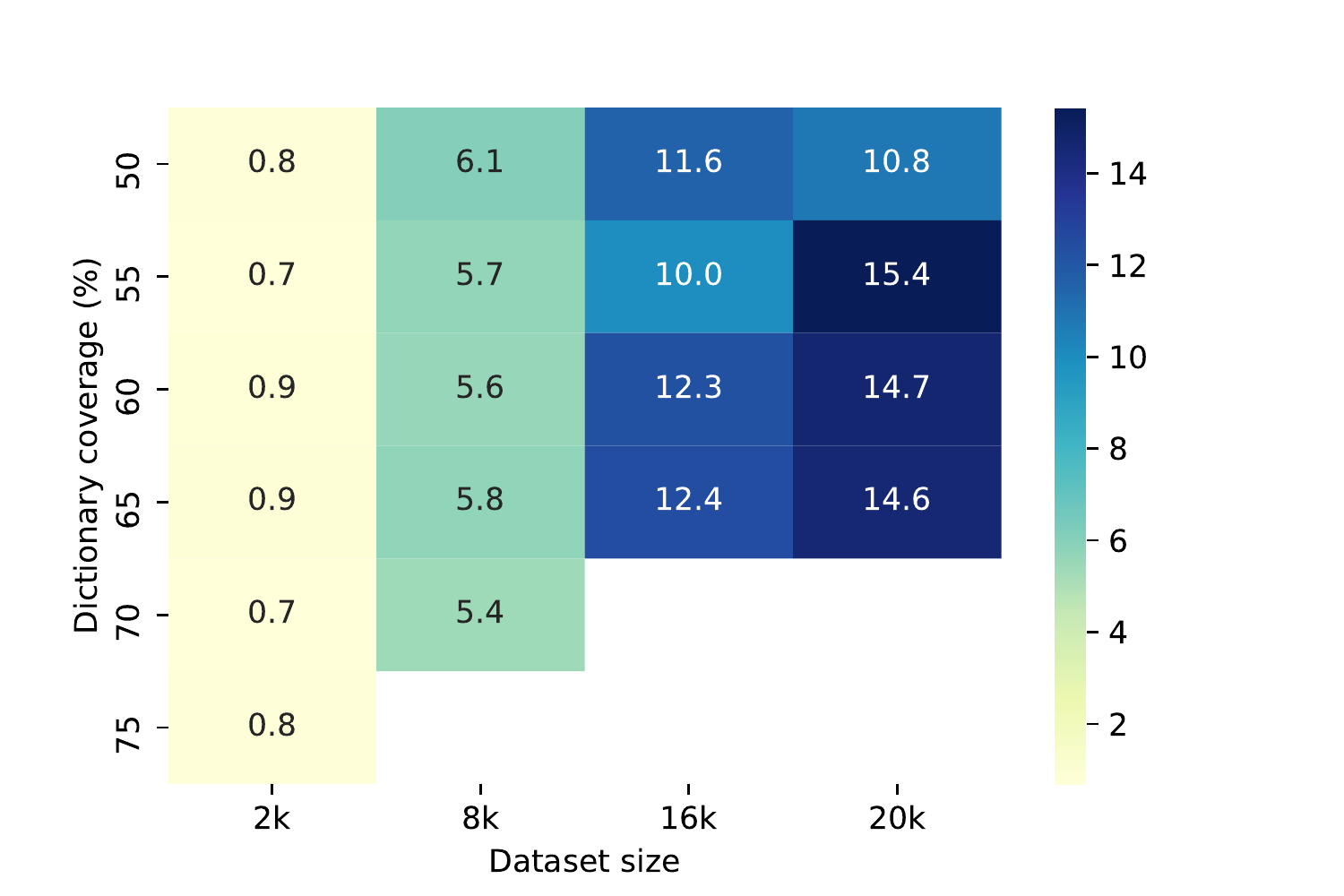}}
\subfigure[Swedish]{\label{}\includegraphics[width=0.32\linewidth]{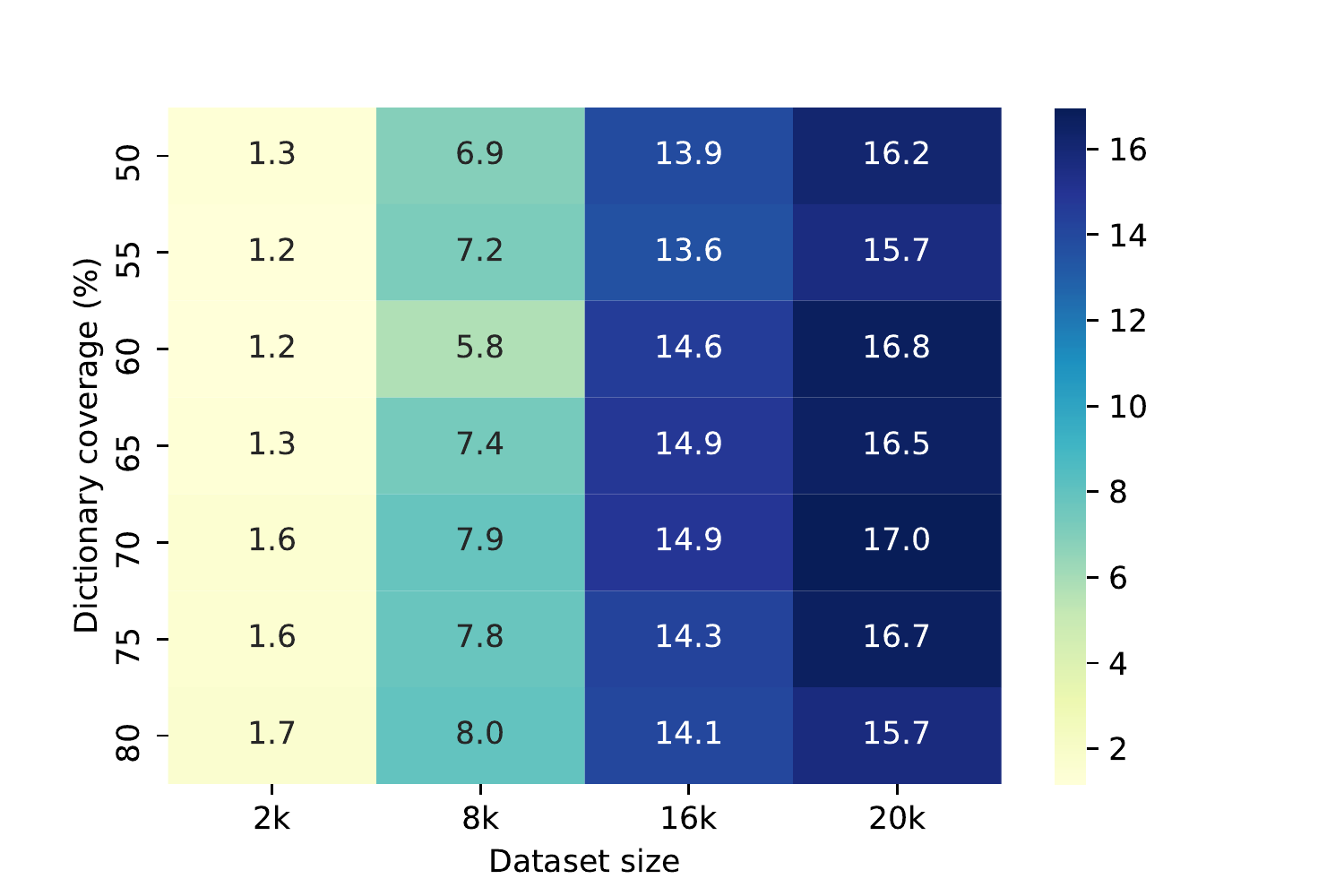}}
\centerline{}
\caption{BLEU Scores for One-to-One translation method. We report the scores on test data with $80\%$ dictionary coverage, as it was maximum in every case.}
\label{fig heatmap}
\end{figure*}
Table \ref{table:results} shows comparison between the baseline (bilingual dictionary based word-to-word translation) and our proposed method. The best scores for each language pair, along with the dictionary coverage are reported in the table. The best result is chosen over the dictionary coverage ($50\%-80\%$), i.e. least percentage of words in each sentence available in the bilingual dictionary, and varied dataset size ($2K-20K$). We report arithmetic mean of scores on $3$ different datasets sampled from the same large data. The scores show a significant increase from simple word-to-word translations ($4.8-8.6$). We performed experiments for three language pairs with simple transformer as well. However, due to very less data, the model seemed to struggle considerably. For Pt-En, Sv-En, and Fr-En language pairs, best scores on $20K$ dataset came to be $1.56$, $1.39$, and $1.37$ respectively. This shows there is a clear advantage of using the proposed method for extremely low resource languages. \par
In Figure \ref{fig heatmap}, we have heat-maps of BLEU scores for different languages calculated over different datasets. The x-axis shows the size of the dataset and the y-axis shows the dictionary percentage. We can have the following observations from the maps, 
\begin{itemize}
    \item \textbf{BLEU VS Dataset size:} The model clearly benefits from increasing the dataset size in an extremely low resource setting. There is a direct correlation between the score and the number of training examples. 
    \item \textbf{BLEU VS Training data dictionary coverage:} Dictionary coverage can be seen as inversely proportional to the amount of noise generated by the untranslated words from the dictionary. The best scores for each column of any map are always somewhere in the middle (except $20K$ dataset of French). We suspect this behavior is linked to finding the correct balance of noise and generalisation. With more noise (less dictionary coverage), the method seems to get more robust, however, it underfits when it is exposed to too many of them.
\end{itemize}



\section{Dict-NMT for many-to-one translation}
A conventional idea for a many-to-one model would involve mapping the source text to a latent representation space which would then further be used by the model to generate the translations. By fixing the target language, we can create a latent representation for any given source language by translating the source text word-to-word into the target language using the bilingual dictionary. This is similar to how we humans translate any foreign language with the help of a bilingual dictionary. \par
We propose a many-to-one translation framework, which, just using a bilingual dictionary, can translate languages that are not present in the training phase- absolute zero-shot translation. Given a test language pair $(S, T)$, we train an NMT model on dictionary-based word-to-word translations of language pairs $\{(S_{i}, T)\}_{i=1}^{n}$, where $S \neq S_{i}$ for $i=1,..,n$. Our goal is to make the model invariant of the source language. We achieve this by adding word-to-word dictionary translations from various languages from different families (Algo $\ref{algo})$. 
\subsection{Experimental Settings}
We perform a comprehensive study on the effect of dataset size, no. of languages, the inclusion of test language family, and dictionary coverage in the test set on the translation quality.
We perform our experiments on European languages with English as the target language. We keep the tokenizer, NMT model, and its hyperparameters similar to the previous experiment's setting. We use Facebook MUSE's bilingual dictionary for this experiment as well. 
\subsubsection{Dataset}
We perform experiments on Europarl v7 parallel corpus, fixing English as our target language. We used languages from three families, namely, Italic (Romanian, Spanish, Portuguese, French, Italian), Slavic (Bulgarian, Czech, Polish, Slovene, Slovak), and Germanic (Danish, Swedish, German, Dutch). We use Romanian, Bulgarian, and Danish as our test languages. We analyze our results on training data, intra and inter combination of the language families with sizes $50K$, $150K$, $500K$, and $1M$. We test our experiments on $600$ sentences. $create\_dataset()$(Algo \ref{algo}) shows how we created training data for our experiment. In our experiments, for a training set $create\_dataset(\{D_{i}\}^{n}_{i=1}$, $\{B_{i}\}_{n}$, p) (Algo \ref{algo}), we take equal number of sentences from all $n$ languages. The case of polysemy is handled the same way as in the previous experiment.
We use the notation "All" for the combination of the above-mentioned languages from all three language families (Italic, Germanic, and Slavic).

\subsection{Results and Analysis}
We present the best scores for three language pairs, Ro-En, Bg-En, and Da-En, in Table \ref{table:results_ULT}. We further compare the scores with word-to-word dictionary translations. We choose the best score over varied training data size ($50K$ - $1M$), Test set dictionary coverage ($0\%$ - $80\%$), and a combination of language families. There is a significant difference in scores of baseline and our proposed method. Because the training sample has no examples from test languages, the resulting score demonstrates the zero-shot property of the proposed method. \par

\begin{table}[h!]
\centering
\begin{tabular}{lrrr}
\hline
\multirow{2}{*}{Language} & \multicolumn{1}{c}{Dict \%}                & \multicolumn{2}{c}{BLEU}              \\ \cline{3-4} 
                          & & Base & w D \\ \cline{1-2}
Ro-En       & $50\%$ &  ${9.4}$  & $\mathbf{28.1}$ \\
Bg-En       & $50\%$ & ${8.2}$   & $\mathbf{15.4}$\\
Da-En     & $50\%$ &  ${7.7}$       & $\mathbf{13.4}$\\ \hline
\end{tabular}
\caption{Results: Best BLEU score for each language. w D is ``with dictionary", i.e. our proposed method, Base is ``dictionary baseline" which is simply word-to-word translation . We get the best scores for "All" dataset (Germanic $+$ Italic $+$ Slavic) with $500K$ parallel sentences. Training set dictionary coverage (Dict $\%$) is given for each corresponding score. The test set of Ro and Bg has atleast $80\%$ dictionary coverage, while Da has $70\%$ (the raw dataset was too little to make a decent test dataset with similar number of samples). }
\label{table:results_ULT}
\end{table}

We perform experiments to test effect of dataset size, inclusion of test language family, and test data dictionary coverage. 
\begin{itemize}
    \item \textbf{BLEU VS Test data dictionary coverage:} From figure \ref{fig:test_dict}, it is evident that the scores increase with the increase in dictionary coverage of test data, i.e., the NMT model gets better assisted with more word-to-word translations in a given sentence.
    \item \textbf{BLEU VS Training set data size:} With increase in data size, the scores increase as well (Table \ref{fig:train_size}). However, it tends to converge on the data size between $500K$ and $1M$.

\begin{figure*}[httb]
    \centering
\subfigure[Bulgarian]{\label{}\includegraphics[width=0.30\linewidth]{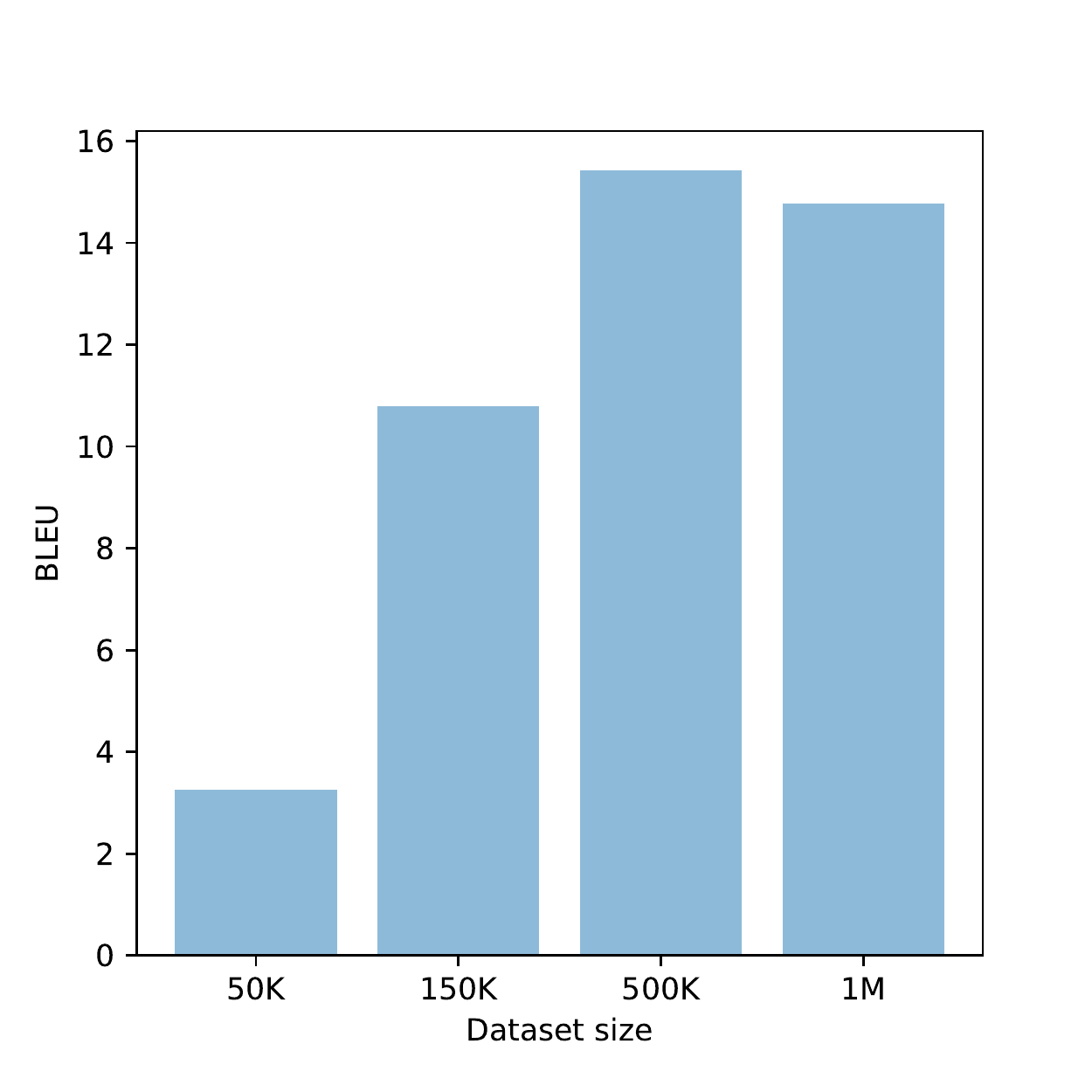}}
\subfigure[Romanian]{\label{}\includegraphics[width=0.30\linewidth]{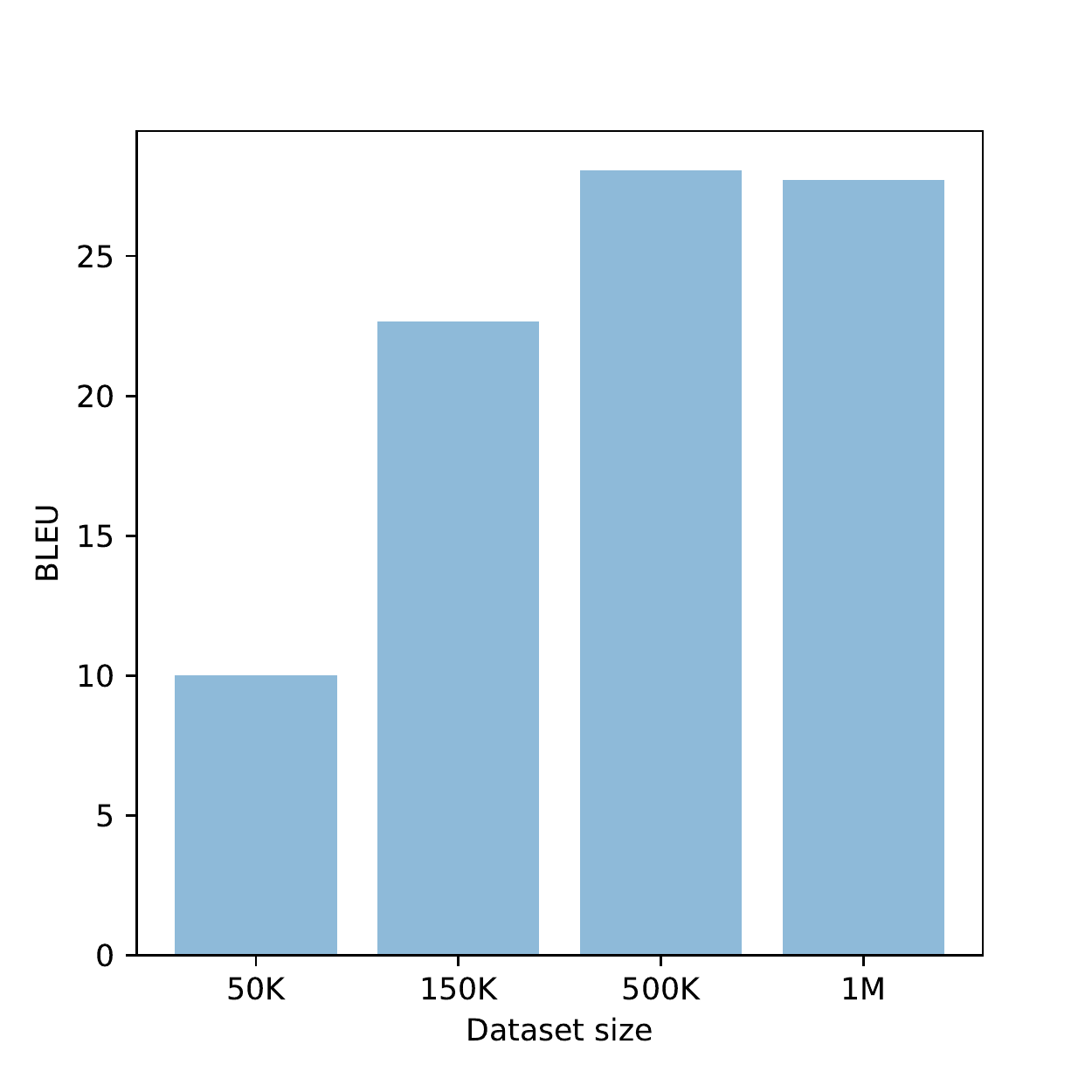}}
\subfigure[Danish]{\label{}\includegraphics[width=0.30\linewidth]{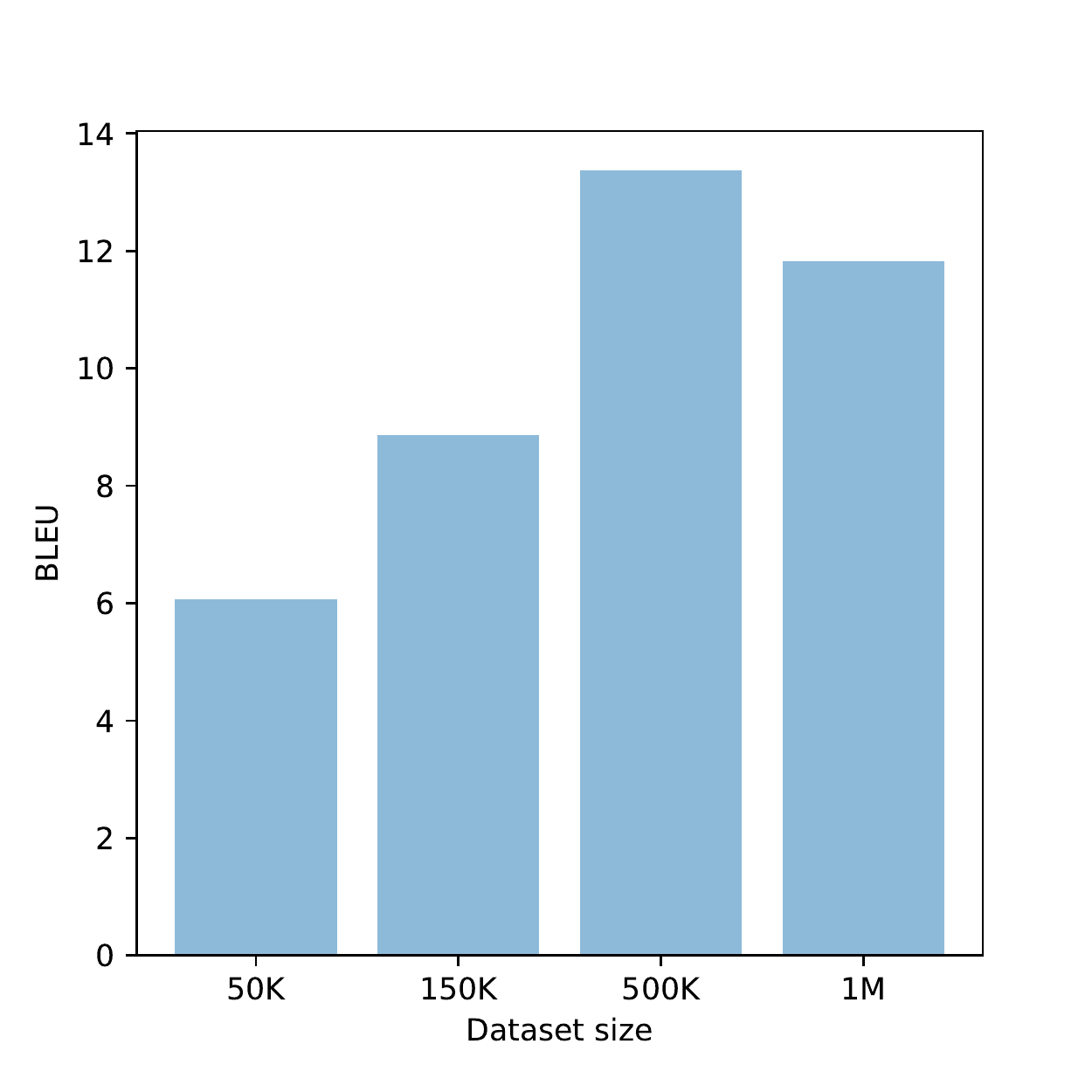}}
\centerline{}
\caption{BLEU VS Training Dataset Size}
\label{fig:train_size}
\end{figure*}

\begin{figure*}[httb]
\centering
\subfigure[Bulgarian]{\label{}\includegraphics[width=0.3\linewidth]{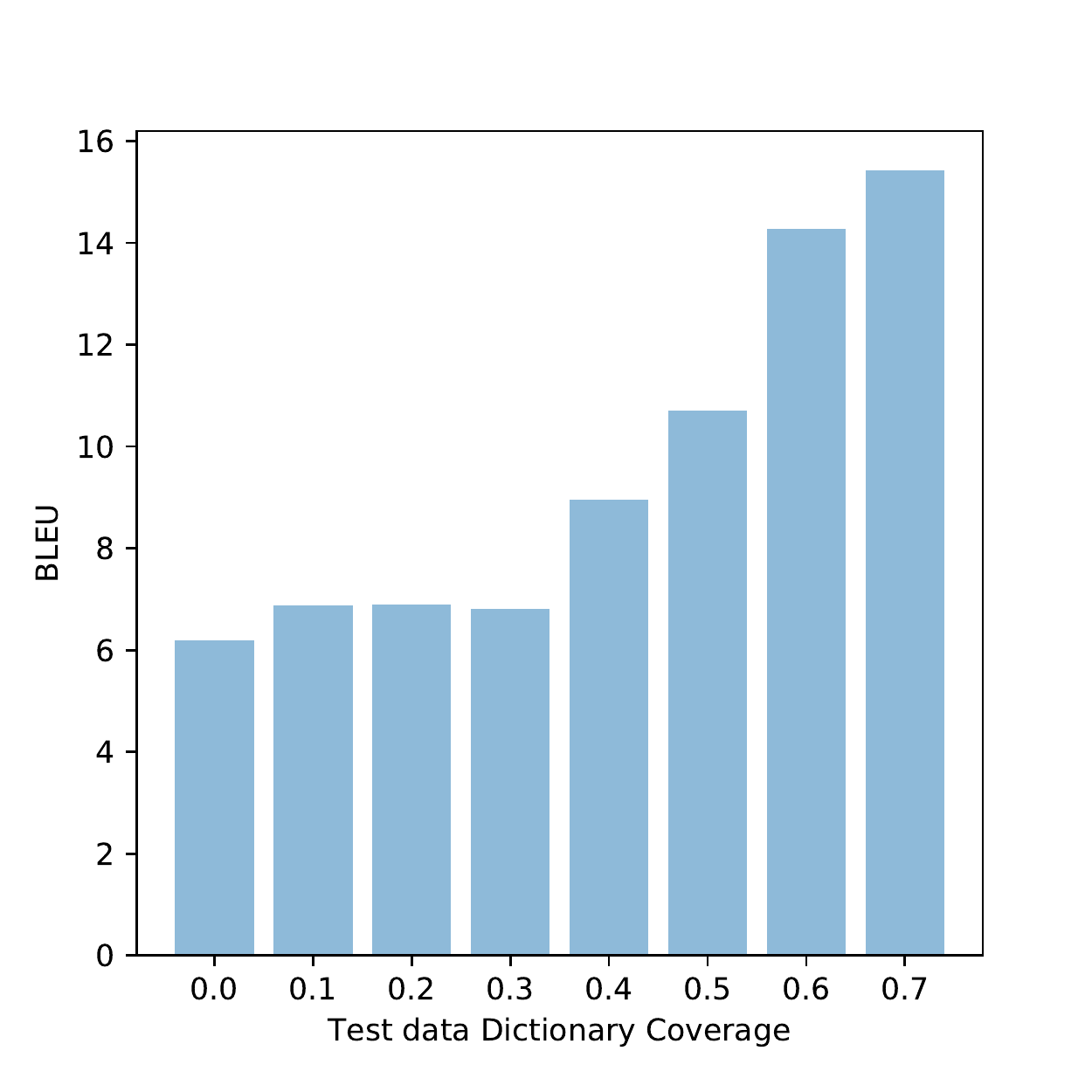}}
\subfigure[Romanian]{\label{}\includegraphics[width=0.3\linewidth]{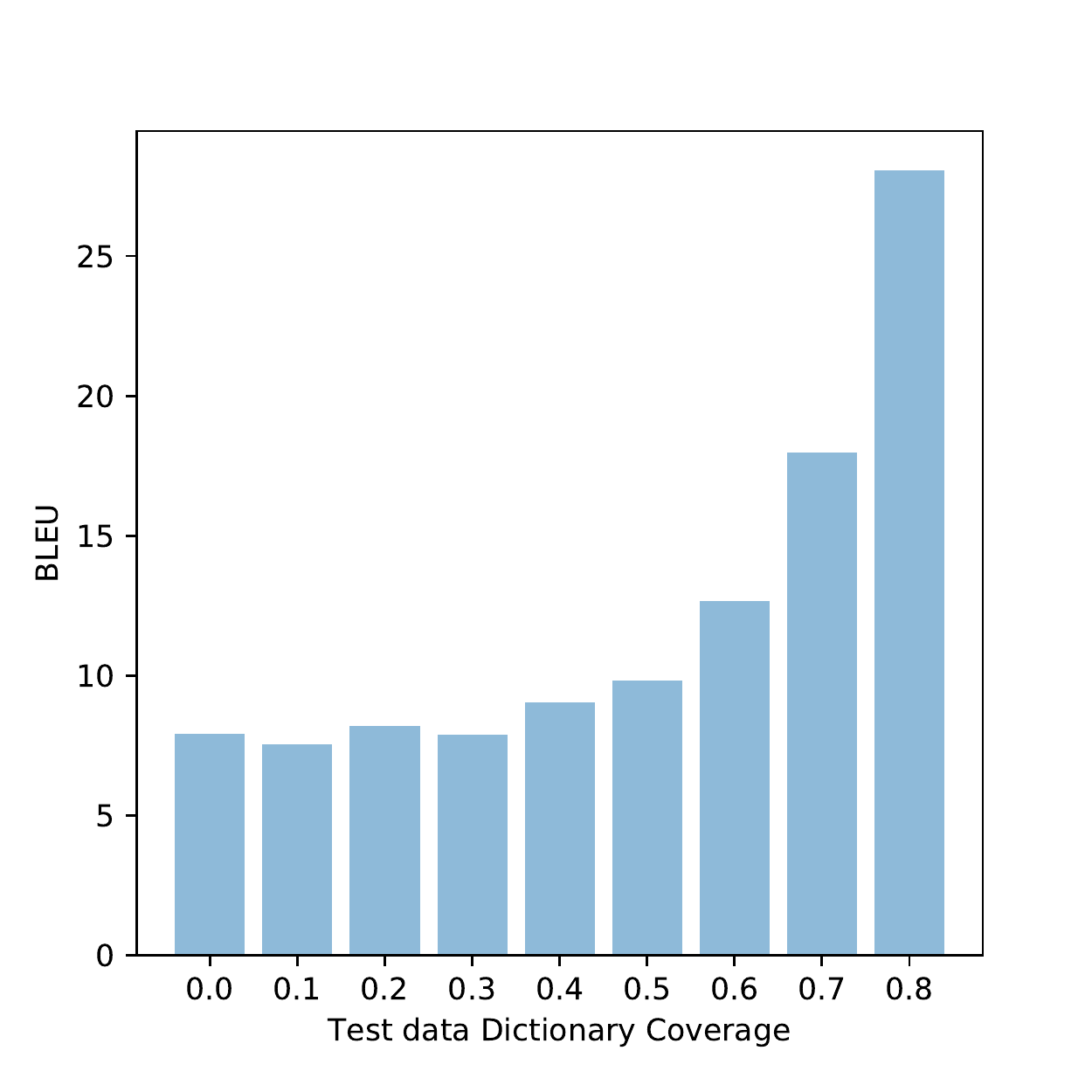}}
\subfigure[Danish]{\label{}\includegraphics[width=0.3\linewidth]{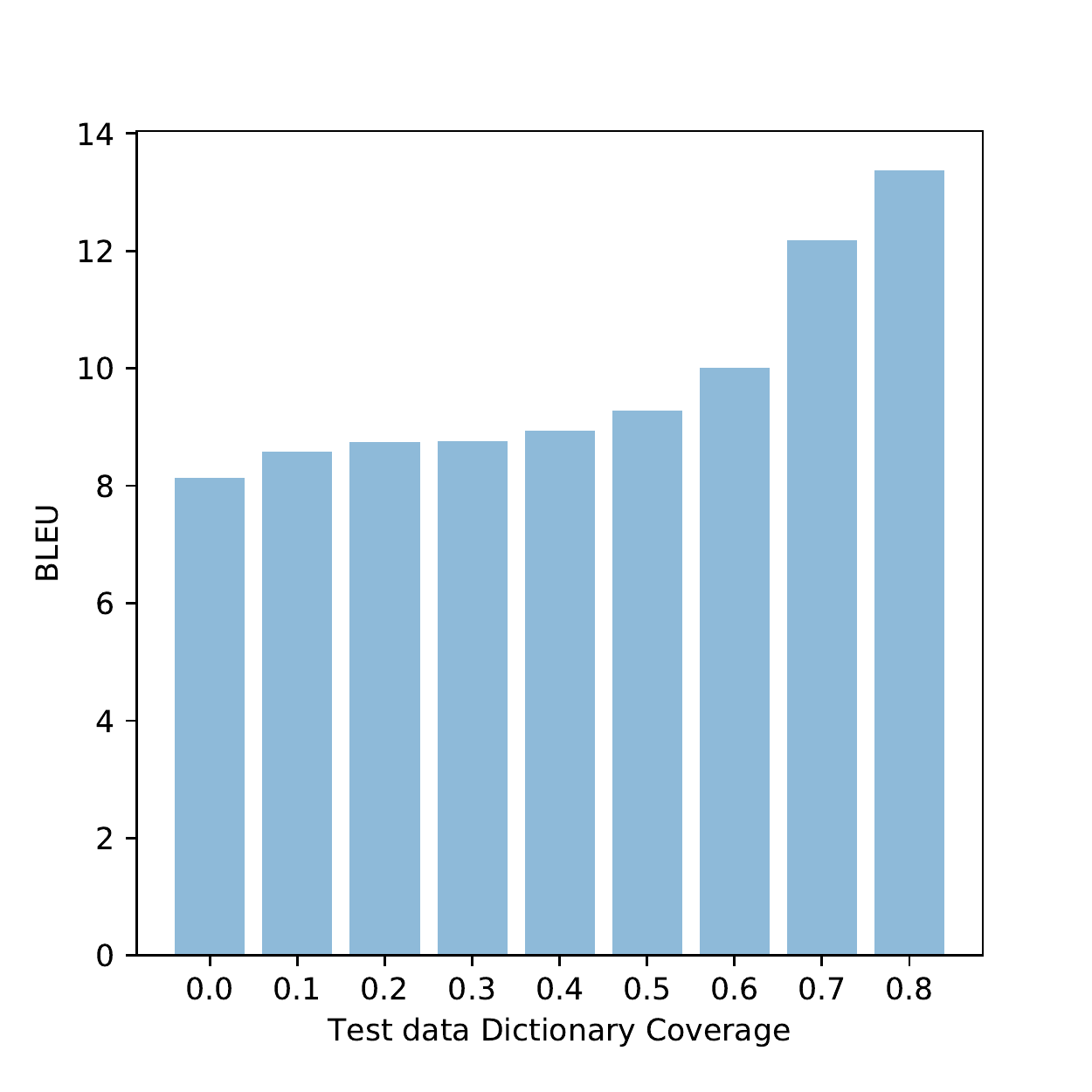}}
\centerline{}
\caption{BLEU VS Test Data Dictionary Coverage. Here, the reported scores are for model trained on "All" dataset with size $500K$}
\label{fig:test_dict}
\end{figure*}

    \item \textbf{Effect of addition of language families:} From Figure \ref{fig:langfam} it can be observed that the score stays the least for model trained just on Germanic family. There is a slight increase in score for Italic family. However, it increases significantly when we start combining language families together. We get the highest score for "All", which is a combination of all three language families. There is a slight decrease in score when we add sentences from two different languages. We suspect less number of parameters of the model to be the reason behind such behaviour. For better generalisation on more number of languages, we believe larger NMT models would be beneficial.
    
\begin{figure}[h!]
\centering
\includegraphics[scale=0.5]{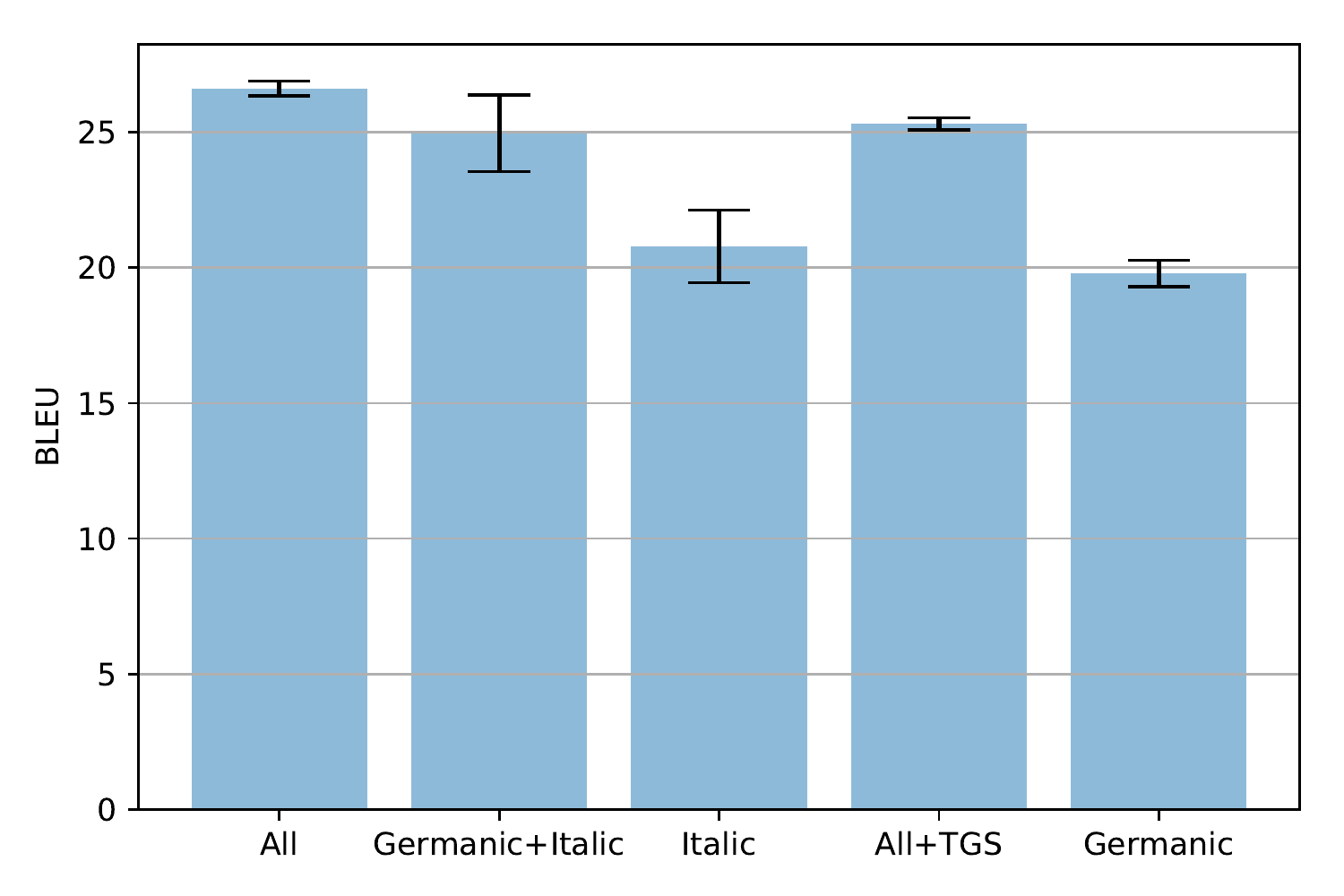}
\caption{BLEU VS Effect of Test Family (For Romanian) (TGS = Turkish + Greeek)}
\label{fig:langfam}
\end{figure}
\end{itemize}




\section{Conclusion}
Using Europarl corpus, we showed that our method of incorporating bilingual dictionaries for NMT tasks could be pretty effective. Given a dictionary, it not only works for languages with extremely low corpus but also for languages with no parallel or monolingual corpus. We analyze the extent of improvement that can be done by varying dictionary percentages and with the range of size of datasets. This work can be extended by blending our method with other state-of-the-art approaches such as back translation and transfer learning. We experiment with our method on only European languages. It would be interesting to evaluate our method's performance for non-European and some syntactically-dissimilar languages.
We believe this work will motivate researchers to explore other possibilities of incorporating bilingual dictionaries for NMT in extremely low resource settings.

\bibliography{dict_nmt}

\bibliographystyle{dict_nmt}

\appendix



\end{document}